\documentclass{article}

\usepackage[nonatbib, final]{nips_2016}

\usepackage[utf8]{inputenc} 
\usepackage[T1]{fontenc}    
\usepackage{hyperref}       
\usepackage{booktabs}       
\usepackage{nicefrac}       
\usepackage{microtype}      

\usepackage[dvipdfmx]{graphicx}
\usepackage{subfigure}
\usepackage[cmex10]{amsmath}
\usepackage{amssymb}
\usepackage{amsfonts}
\usepackage{amsthm}
\usepackage{url}
\usepackage{bm}
\usepackage{multirow}
\usepackage{color}

\newcommand{\argmin}{\mathop{\mathrm{argmin}}\limits}

\newcommand{\supp}{\mathrm{supp}}

\usepackage{CJKutf8}
\usepackage{ifthen}
\newcommand{\COMM}[2]{{
\begin{CJK}{UTF8}{ipxm}
\ifthenelse{\equal{#1}{SH}}{\color{blue}}{
\ifthenelse{\equal{#1}{TM}}{\color{red}}{
\ifthenelse{\equal{#1}{AA}}{\color{cyan}}{
\ifthenelse{\equal{#1}{BB}}{\color{magenta}}}}}
[#1: #2]
\end{CJK}
}}

\title{Finding Alternate Features in Lasso}

\author{
Satoshi Hara\\
National Institute of Informatics, Japan \\
JST, ERATO, \\
Kawarabayashi Large Graph Project \\
\texttt{satohara@nii.ac.jp} \\
\And
Takanori Maehara\\
Shizuoka University, Japan \\
JST, ERATO, \\
Kawarabayashi Large Graph Project \\
\texttt{maehara.takanori@shizuoka.ac.jp} \\
}

\begin{document}

\maketitle

\begin{abstract}
We propose a method for finding alternate features missing in the Lasso optimal solution.
In ordinary Lasso problem, one global optimum is obtained and the resulting features are interpreted as task-relevant features.
However, this can overlook possibly relevant features not selected by the Lasso.
With the proposed method, we can provide not only the Lasso optimal solution but also possible alternate features to the Lasso solution.
We show that such alternate features can be computed efficiently by avoiding redundant computations.
We also demonstrate how the proposed method works in the 20 newsgroup data, which shows that reasonable features are found as alternate features.
\end{abstract}

\section{Introduction}
\label{sec:intro}

Feature selection is a procedure that selects a subset of relevant features (i.e., variables) for model construction.
It helps users to understand which features are contributing to the model.
Hence, it is a most basic approach for model interpretation in machine learning.
It is important to note that the quality of selected features heavily affects the user's trust on the resulting model.
It is common that domain experts have some prior knowledge about data which features are important for the proper task.
If such features are not selected by the feature selection, the experts will not trust the model because it does not agree with their intuition.
If the model is not trusted by the experts, the model will be never used even if it may perform well in practice.
It is therefore important for feature selection methods to meet the user's demand by not missing important features.

One of the most common feature selection methods is Lasso~\cite{tibshirani1996regression,chen2001atomic}.
We consider a prediction problem with $n$ observations and $p$ predictors.
Here, we have a response vector $y \in \mathcal{Y}^n$ and a predictor matrix~$X \in \mathbb{R}^{n \times p}$ where $\mathcal{Y}$ is the domain of the response (e.g., $\mathcal{Y}=\mathbb{R}$ for regression, and $\mathcal{Y}=\{-1, 1\}$ for classification). 
In the Lasso problem, we seek $\beta \in \mathbb{R}^p$ that minimizes $\ell_1$-regularized objective function:
\begin{align}
  L(\beta) := f(X\beta, y) + \rho \| \beta \|_1
  \label{eq:lasso}
\end{align}
where $f: \mathbb{R}^n \times \mathcal{Y}^n \to \mathbb{R}_{\ge 0}$ is a loss function, and $\rho \in \mathbb{R}_{\ge 0}$ is a regularization parameter.
The optimal solution $\beta^* \in \mathbb{R}^p$ to \eqref{eq:lasso} is usually sparse;
therefore, we can extract a set of features as the support of the optimal solution, $\supp(\beta^*)=\{i : |\beta^*_i| > 0\}$.

In ordinary Lasso problem, one global optimum $\beta^*$ is obtained and the resulting features are interpreted as task-relevant features.
However, this can overlook possibly relevant features not selected by the Lasso.
Indeed, Lasso can recover \emph{true} features only under some limited conditions~\cite{knight2000asymptotics,wainwright2009sharp}.
We are therefore in a risk of missing important features if the conditions are not met.
One particular example of the Lasso failure is when two features $x_i$ and $x_j$ are highly correlated.
In such a situation, Lasso tends to select only one of these two features (e.g., $\beta_i^* \neq 0$ while $\beta_j^* = 0$).
That is, we may overlook one of these two features although both of them may contribute to the task equally.

In this study, we propose a method for finding task-relevant features missing in the Lasso optimal solution $\beta^*$.
In particular, we seek for whether there are any alternate feature $x_j$ that can be replaced with a feature $x_i$ selected by the Lasso.
With this procedure, we can provide not only the Lasso optimal solution but also alternate features missed in the solution to the users.
Even if some important features are missed in the Lasso optimal solution, such features are likely to be selected as a part of alternate features.
Hence, the user's trust on the resulting model will be greatly improved because they can find out that important features are actually not missed but replaced with some other features.
Moreover, the users can customize the model based on the information about alternate features; one can remove the feature $x_i$ selected by the Lasso and add an alternate feature $x_j$ to the model instead so that the model to agree with the user's background knowledge.

\section{Finding Alternate Features}
\label{sec:method}

Given the Lasso optimal solution $\beta^*$, we seek for whether there are any alternate feature $x_j$ with $\beta^*_j=0$ that can be replaced with a feature $x_i$ selected by the Lasso (i.e., $\beta^*_i \neq 0$).
We solve this problem by optimizing $\beta_j$ in (\ref{eq:lasso}) while fixing as $\beta_i = 0$ and $\beta_k = \beta^*_k \, (k \neq i, j)$.
The optimization problem can be expressed as
\begin{align}
	\beta^{(i)}_j = \argmin_{\beta_j} f(z^{(i)} + X_j \beta_j, y) + \rho | \beta_j | ,
	\label{eq:sublasso}
\end{align}
where $X_j$ denotes the $j$-th column of $X$ and $z^{(i)} = \sum_{k \neq i} X_k \beta^*_k$.
If $\beta^{(i)}_j \neq 0$, the feature $x_j$ can be an alternative of $x_i$.
We note that the problem (\ref{eq:sublasso}) is a univariate optimization problem, and can be solved easily, e.g., by using the proximal gradient method~\cite{boyd2004convex}.

To find out all possible $(i, j)$-pairs, we basically need to solve the problem (\ref{eq:sublasso}) for all $i \in \supp(\beta^*)$ and $j \in \supp(\beta^*)^c$.
Here, we show that we actually need to solve the problem (\ref{eq:sublasso}) only on a fraction of $j$ instead of all $j \in \supp(\beta^*)^c$.
This is because we can check that $\beta^{(i)}_j = 0$ \emph{without} solving the problem (\ref{eq:sublasso}) from the optimality condition; $\beta^{(i)}_j = 0$ holds when $|X_j^\top \nabla f(z^{(i)}, y)| \le \rho$ where $\nabla f$ is the derivative of $f$ over the first element.
Hence, we need to solve the problem (\ref{eq:sublasso}) only for $j \in \supp(\beta^*)^c$ with $|X_j^\top \nabla f(z^{(i)}, y)| > \rho$.

\paragraph{Scoring Alternate Features}
By using the proposed method, we can find a set of alternate features of $x_i$, namely $\{j : \beta_j^{(i)} \neq 0\}$.
Among several alternate features, it is of great interest to find alternate features that closely relates to the original feature $x_i$.
Here, we propose a scoring method for each alternate feature so that we can find such interesting features.
The proposed scoring method is based on the Lasso objective function $L(\beta)$.
Let $\beta^*$ be the Lasso optimal solution, and $\beta^{i \rightarrow j}$ be the alterante solution defined by $\beta^{i \rightarrow j}_i = 0$, $\beta^{i \rightarrow j}_j = \beta_j^{(i)}$, and $\beta^{i \rightarrow j}_k = \beta_k^* \, (k \neq i, j)$.
The relevance of the alternate feature $x_j$ to the original feature $x_i$ can be measured by using the increase of the objective function value, ${\rm score}(x_i \rightarrow x_j) = L(\beta^{i \rightarrow j}) - L(\beta^*)$.
If the alternate feature $x_j$ is almost identical to the original feature $x_i$, it is likely that the objective function value $L(\beta^{i \rightarrow j})$ is almost the same as $L(\beta^*)$, which results in small ${\rm score}(x_i \rightarrow x_j)$.
Hence, we can use this score to order alternate features so that we can find out particularly related features.

\section{Experimental Results on 20 Newsgroups Data}
\label{sec:exp}

The 20 Newsgroups\footnote{\url{http://qwone.com/~jason/20Newsgroups/}} is a dataset for text categorization.
In this experiment, we tried to find discriminative words between the two categories\footnote{The experiment codes are available at \url{https://github.com/sato9hara/LassoVariants}}.
In the first task, we considered categories \texttt{ibm.pc.hardware} and \texttt{mac.hardware}, and in the second task, we considered \texttt{sci.med} and \texttt{sci.space}.
As a feature vector $x$, we used tf-idf weighted bag-of-words expression, with stop words and some common verbs removed.
See \tablename~\ref{tab:data} for the detail of the datasets.
The tasks were to find discriminative words that were relevant to classification.

Because the task was binary classification between the two categories, we used Lasso logistic regression~\cite{lee2006efficient}.
The logistic loss function is defined by $f(z, y) := \sum_{m=1}^n \log (\exp{(-y_m z_m)} + 1)$.

In the experiment, we set the regularization parameter as $\rho=0.001n$, and derived the Lasso optimal solution $\beta^*$.
As the optimal solution $\beta^*$, 39 words and 31 words were selected as relevant for classification in the first task and the second task, respectively.
To find out all $(i, j)$-pairs, in the first task, the naive approach required solving the problem (\ref{eq:sublasso}) for $39 \times 11,609 \approx 450,000$ times, while, by using the proposed checking method, this number was reduced to only $53$ times which was almost 10,000 times smaller than the naive approach.

\begin{figure}[tb]
	\centering
	\includegraphics[width=0.9\textwidth]{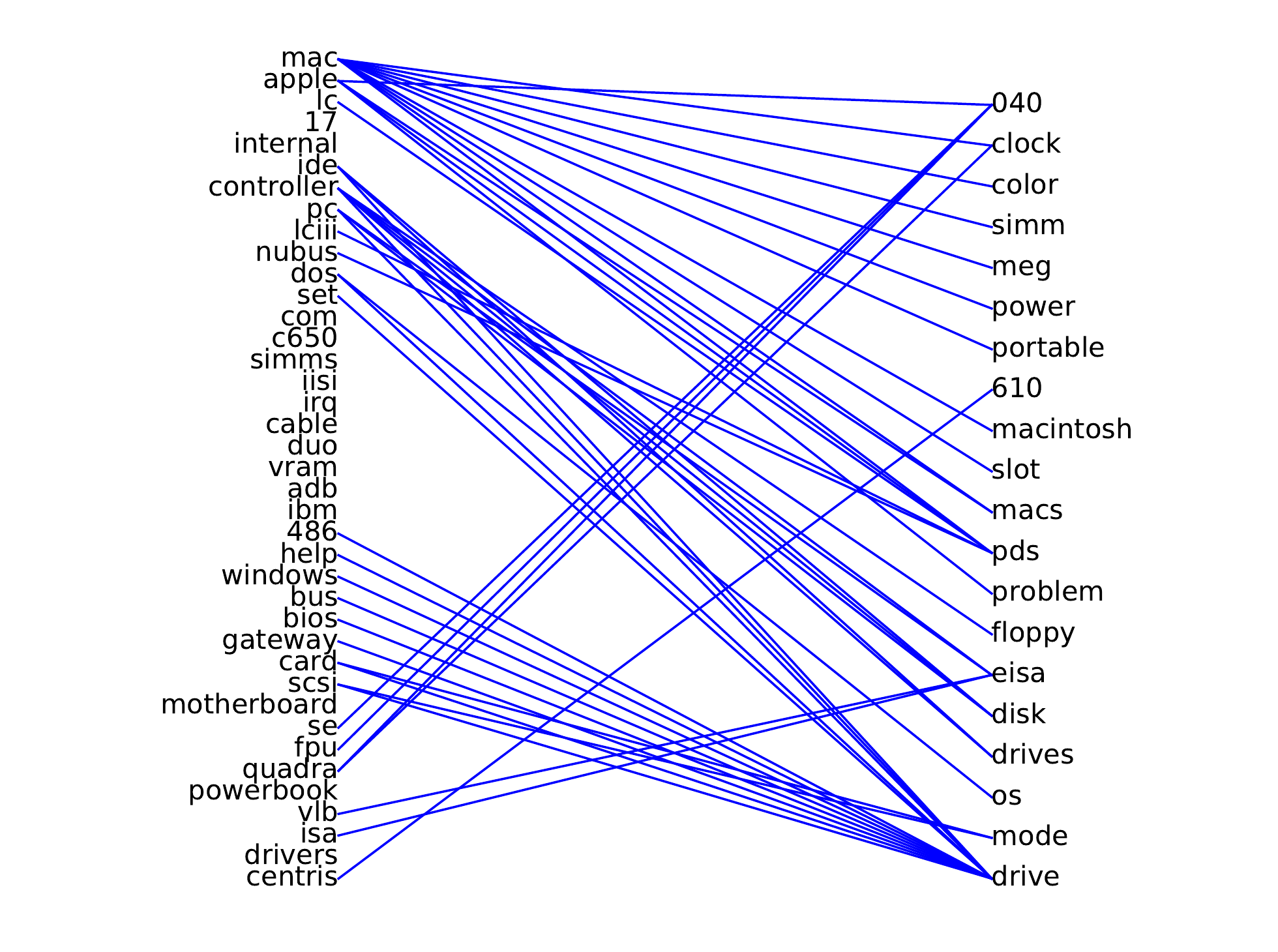}
	\caption{[Found feature pairs in \texttt{ibm.pc.hardware} vs \texttt{mac.hardware}] Left: words in the Lasso optimal solution, Right: alternate words of the left words connected by the edges.}
	\label{fig:news20_1}
\end{figure}

\begin{figure}[tb]
	\centering
	\includegraphics[width=0.9\textwidth]{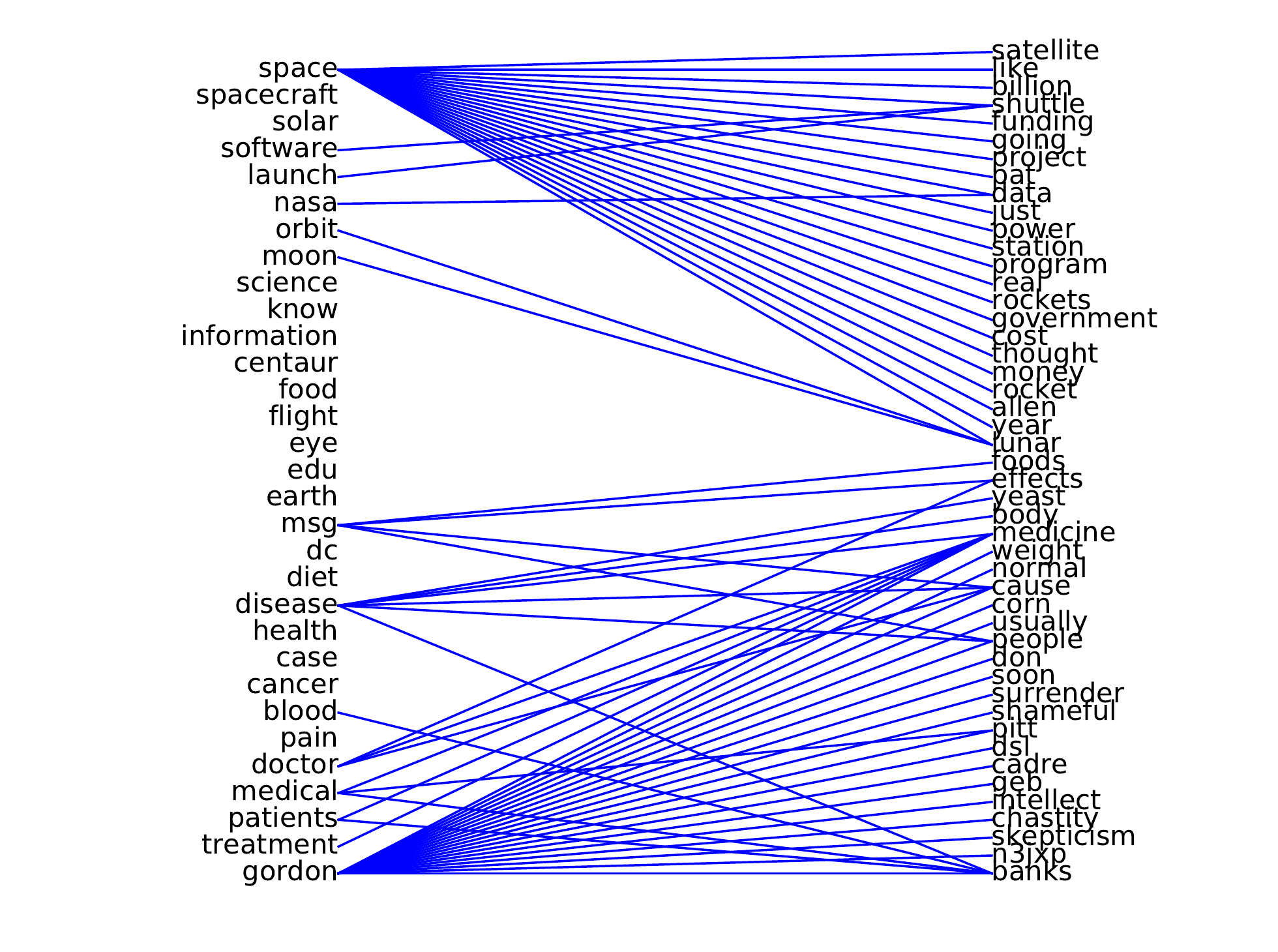}
	\caption{[Found feature pairs in \texttt{sci.med} vs \texttt{sci.space}] Left: words in the Lasso optimal solution, Right: alternate words of the left words connected by the edges.}
	\label{fig:news20_2}
\end{figure}

The found feature pairs are shown in \figurename~\ref{fig:news20_1} and \ref{fig:news20_2}.
From \figurename~\ref{fig:news20_1}, we can find several interesting feature pairs.
For instance, the alternate word \emph{drive} is paired with many words such as \emph{windows}, \emph{bus}, and \emph{bios}, which are all related to the Windows machine (i.e., the words related to \texttt{ibm.pc.hardware}).
Another interesting finding is the pair \emph{centris} and \emph{610}, both of which are from the Mac's product name Centris 610 (i.e., the words related to \texttt{mac.hardware}).
Not limited to the examples above, but the found pairs seem to be quite reasonable.
Hence, providing these found alternate words together with the Lasso optimal solution will make the resulting model more trustful to the users compared to just providing the Lasso optimal solution.

\begin{table}[t]
	\centering
	\caption{20 Newsgroups Data: Each feature corresponds to each word appearing in the texts.}
	\label{tab:data}
	\begin{tabular}{c|cc}
		& \# of features $p$ & \# of observations $n$ \\ \hline
		\texttt{ibm.pc.hardware} vs \texttt{mac.hardware} & 11,648 & 1,168 \\
		\texttt{sci.med} vs \texttt{sci.space} & 21,369 & 1,187
	\end{tabular}
	\vspace{-8pt}
\end{table}

\figurename~\ref{fig:news20_2} shows that two words \emph{space} and \emph{gordon} are connected with many alternate words.
The word \emph{space} and its alternate words such as \emph{shuttle} and \emph{satellite} are convincing as these words are all related with the category \texttt{sci.space}.
On the other hand, the word \emph{gordon} and its alternate words such as \emph{banks}, \emph{skepticism}, and \emph{shameful} seem not to be relevant to neither of the categories \texttt{sci.med} nor \texttt{sci.space}.
These words actually come from the frequently appearing signature in \texttt{sci.med}: \emph{Gordon Banks  N3JXP, geb@cadre.dsl.pitt.edu, Skepticism is the chastity of the intellect, and it is shameful to surrender it too soon.}
That is, the model is trained to find this signature and classify the text into the category \texttt{sci.med}.
In practice, this model is not preferable as the model is too specialized to this specific task; the trained model may perform poorly for the texts that do not include this signature.

\begin{figure}[tb]
	\centering
	\subfigure{\includegraphics[width=0.49\textwidth]{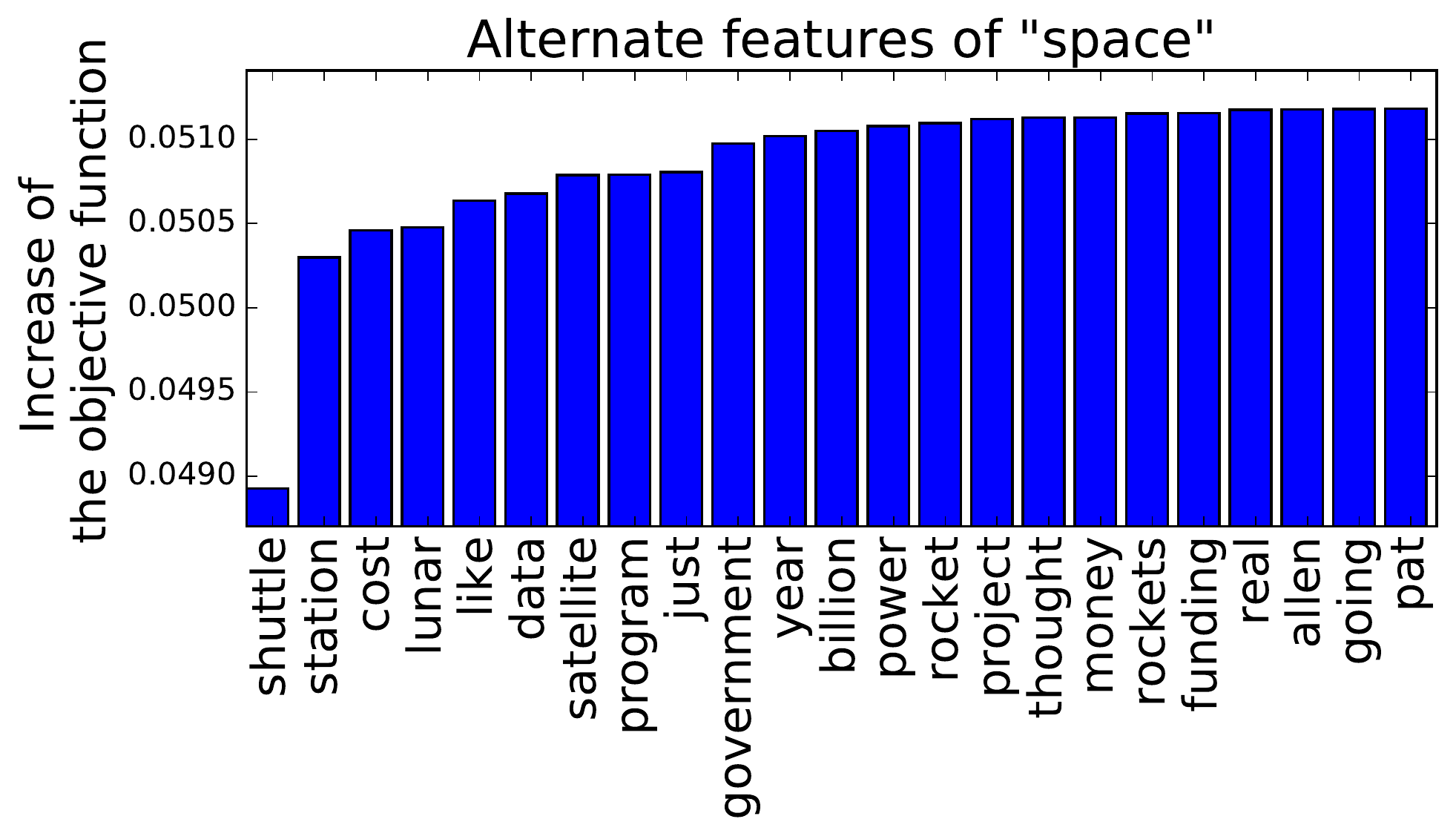}}
	\subfigure{\includegraphics[width=0.49\textwidth]{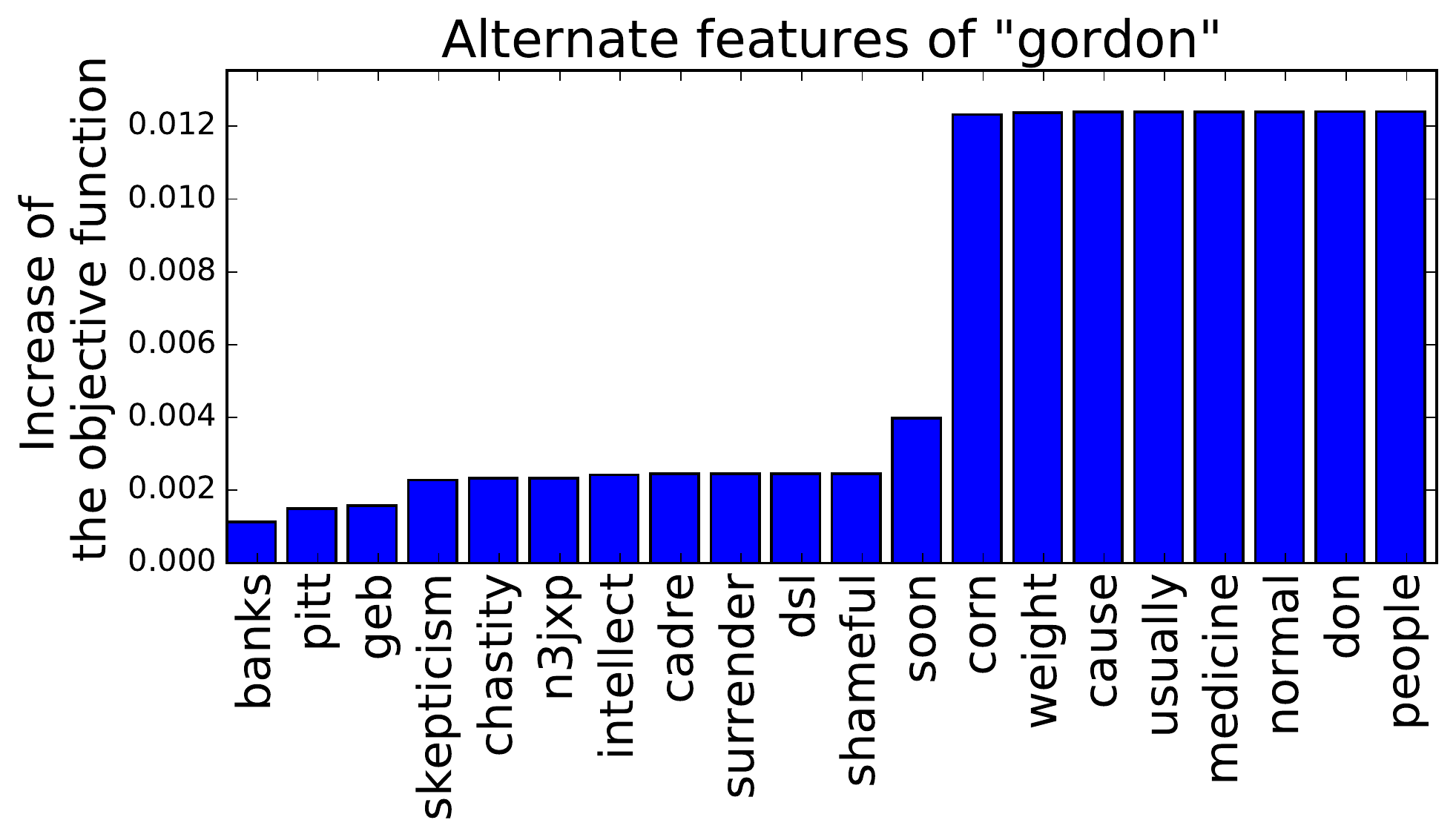}}
	\caption{Alternate features of \textit{space} and \textit{gordon} ordered by the increase of the objective function value ${\rm score}(x_i \rightarrow x_j) = L(\beta^{i \rightarrow j}) - L(\beta^*)$.}
	\label{fig:news20_score}
\end{figure}

To see the result in the second task in detail, we scored alternate features of \textit{space} and \textit{gordon} as shown in \figurename~\ref{fig:news20_score}.
Here, we can find two interesting results.
First, the word \textit{space} is particularly closely related with \textit{shuttle} among several alternate words.
This is a reasonable result because the existence of the word \textit{shuttle} in the text implies that the text's category is \texttt{sci.space} rather than \texttt{sci.med}.
Second, the word \textit{gordon} is particularly closely related with the words appearing in the signature.
Indeed, all of the top 12 words (from \textit{banks} to \textit{soon} in the figure) come from the signature.
This result presents one particular use case of the proposed method.
By scoring several alternate features, we can find out seemingly not preferable features and remove them in the training phase so that the resulting model to agree with our intuition.


\section{Conclusion}
\label{sec:concl}

We proposed a method for finding alternate features missing in the Lasso optimal solution.
With the proposed method, we can provide not only the Lasso optimal solution but also alternate features to the users.
We believe that providing such surrogate information helps the users to interpret the model and encourage them to use the model in practice.

There remains several open issues.
First, we need to make it easy for the users to check all the found feature pairs.
The bipartite graph expression as in \figurename~\ref{fig:news20_1} and \ref{fig:news20_2} would be one possible approach, although it may become too complicated when there are more features.
We think the bipartite graph clustering will be a promising method to simplify the graph.
Second, in the current study we considered replacing only one feature with another feature.
This framework can be naturally extended to replacing multiple features.
Developing an efficient algorithm for the generalized problem remains open.
Finally, there remains a fundamental question that in what circumstances we can find \emph{true} alternate features.
To provide a reliable surrogate information to the users, we need to study the theoretical aspects of the proposed method.


\bibliographystyle{unsrt}
\bibliography{nipsw16}

\end{document}